%% file: main.tex
\documentclass[runningheads]{llncs}
\usepackage{booktabs}
\usepackage{amsmath}
\usepackage[T1]{fontenc}
\usepackage{graphicx}
\usepackage{xcolor}

\newcommand{\ask}[1]{}
\newcommand{\revise}[1]{\textcolor{black}{#1}}
\newcommand{\revised}[1]{{\color{black}#1}}

\begin{document}
\title{D$^2$Styler: Advancing Arbitrary Style Transfer with Discrete Diffusion Methods }
\titlerunning{Paper Title}
%
\author{Onkar Susladkar\inst{1}\orcidID{0000-0003-4511-1858} \and
Gayatri Deshmukh\inst{1}\orcidID{0000-0001-6442-0782} \and
Sparsh Mittal\inst{2}\orcidID{0000-0002-2908-993X}
\and
Parth Shastri \inst{1}\orcidID{0000-0002-1581-984X}
}
\authorrunning{Authors}
%
\institute{$^1$Independent Researcher, 
$^2$Indian Institute of Technology, Roorkee, India
\email{\{onkarsus13,dgayatri9850\}@gmail.com,\\sparsh.mittal@ece.iitr.ac.in,shastripp18.extc@coeptech.ac.in }\\
}
\maketitle              

\input{00abstract}
\input{05introduction}

\input{10related_work}

\input{20methodology}

\input{30experimental}

\input{40quantResults}

\input{50qualResults}

\input{55qualResultsSupp}

\input{60versatilityResults}
\input{80ablation}
\input{99conclusion}

\bibliographystyle{splncs04}
\bibliography{citations}

\end{document}

%% file: 00abstract.tex
\begin{abstract}
    In image processing, one of the most challenging tasks is to render an image's semantic meaning using a variety of artistic approaches. 
    Existing techniques for arbitrary style transfer (AST) frequently experience mode-collapse, over-stylization, or under-stylization due to a disparity between the style and content images. We propose a novel framework called D$^2$Styler (Discrete Diffusion Styler) that leverages the discrete representational capability of VQ-GANs and the advantages of discrete diffusion, including stable training and avoidance of mode collapse. Our method uses Adaptive Instance Normalization (AdaIN) features as a context guide for the reverse diffusion process. 
    This makes it easy to move features from the style image to the content image without bias. The proposed method substantially enhances the visual quality of style-transferred images, allowing the combination of content and style in a visually appealing manner. 
    We take style images from the WikiArt dataset and content images from the COCO dataset. 
    Experimental results demonstrate that D$^2$Styler produces high-quality style-transferred images and outperforms  \textbf{twelve}  existing methods on nearly all the metrics. The qualitative results and ablation studies provide further insights into the efficacy of our technique.  The code is available at \url{https://github.com/Onkarsus13/D2Styler}.  

\keywords{Neural Style Transfer  \and Vector Quantization \and Latent Diffusion}
\end{abstract}

%% file: 05introduction.tex
\section{Introduction}

    \begin{figure}
    \centering
\includegraphics[width=\textwidth]{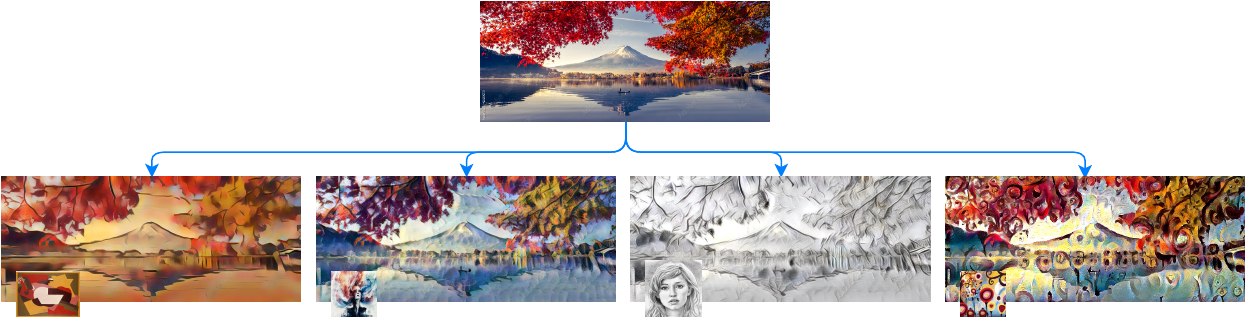}   
\caption{Results from $D^{2}Styler$, our  proposed method. The content image (the one on the top of the tree) is converted to different stylized versions based on the corresponding style image (shown in the inset). }\label{fig:teaser}
    \end{figure}%

    Style transfer (ST) is essential in editing images and generating new artistic images. 
    Given a content image and a style image, Style transfer (ST)  synthesizes a new image by transferring the style from the style image while preserving the substance from the content image.
    Neural style transfer (NST) has been studied extensively in recent years \cite{gatys2016image}.  
    NST seeks to learn how humans perceive images, as transferring style without significantly altering the semantic content in the target image requires a single network to disentangle the style and the content. 
    After the introduction of pioneering work by Gatys et al. \cite{gatys2016image}, many works have proposed improvements in use of a  single feed-forward method \cite{johnson2016perceptual}, loss function, use of regularization, and normalization techniques.

    While GANs have been used to perform NST \cite{kim2019ugatit}, GANs are challenging to train and offer no control over the style and content of the output image. Researchers have used standalone flow-based models and GAN plus flow-based models for image generation.  These models allow control over the output image attributes \cite{fan2022styleflow,an2021artflow}. Recently, diffusion models have become popular for image generation \cite{nichol2021improved,rombach2022high,saharia2022palette}. These models allow generating new images based on text prompts, image in-painting, and conditional image-to-image translation. These models also allow arbitrary style transfer by giving an image and a text prompt specifying the style we need to transfer. However, these models face a strict trade-off between style transfer and content preservation. 
       
    We present $D^2$Styler, a technique to perform arbitrary style transfer for a given content image and a style image. Figure \ref{fig:teaser} illustrates $D^2$Styler output.  $D^2$Styler uses a pretrained VQ-GAN encoder \cite{esser2021taming} to encode the content and style images. Then, it models their combined latent space by a conditional diffusion model. This diffusion model is conditioned on the features extracted by matching the statistics of the content and style images. To achieve this, an AdaIN \cite{huang2017arbitrary} layer with a pre-trained convolutional encoder network is used. 
    Our key idea is that this approach provides a context for the diffusion decoder to predict the masked inputs correctly. The resultant learned latent code is passed through the VQ-GAN decoder to obtain the style-transferred image. 
    Our key contributions are:

    1. We introduce a pioneering approach that combines discrete diffusion with AdaIN, uniquely addressing the prevalent issues of mode-collapse and over-stylization in existing style transfer methods. This integration not only stabilizes the training process but also ensures the preservation of content integrity while applying diverse artistic styles, a critical improvement over prior methodologies.

     2. We propose using AdaIN \cite{huang2017arbitrary} features to guide the diffusion decoder, conditioning the model on matched statistics between style and content features. This approach enables more precise control over the style transfer process, ensuring that the output closely aligns with the desired stylistic attributes while maintaining the integrity of the content.

    3. We propose a new loss function
    ($L_{feature}$) that drives the output image to match the features from the AdaIN \cite{huang2017arbitrary} layer. This loss, coupled with style and content losses, makes the model generate plausible stylized images.

    4. A rigorous evaluation on multiple benchmarks show that D$^2$Styler  outperforms \revised{twelve} established style transfer techniques \cite{gatys2016image,kotovenko2021rethinking,fan2022styleflow,zhang2021image,lee2022cartoon,deng2022stytr2,zhang2022domain,zhang2023inversion,liu2021adaattn,rombach2022high,peebles2023scalable}  on key metrics such as SSIM and LPIPS. This not only demonstrates the effectiveness of our method but also highlights its robustness across various artistic and content scenarios. $D^2$Styler has 78M parameters.

    5. We have shown that D$^2$Styler can effectively apply multiple styles to a single content image (Section \ref{sec:versatility}). This capability significantly expands artistic possibilities, allowing for the creative blending of diverse styles within a single output, which is especially valuable in complex digital art and design projects.

    6. D$^2$Styler is designed to achieve high-quality results in fewer diffusion steps than conventional diffusion architectures and auto-regressive methods \cite{ho2020denoising} (Section \ref{sec:ablation}). This significantly reduces computational costs and improves the feasibility of deploying style transfer in real-time scenarios, thus addressing a major bottleneck in the adoption of NST technologies.

    7. Ablation studies illustrate the contributions of each component of D$^2$Styler, providing clear evidence to the community on the effectiveness of methodology.

%% file: 10related_work.tex
\section{Related Work}\label{sec:relatedWork}              
    \textbf{Neural Style Transfer (NST):}         
        NST has been extensively studied in non-photorealistic rendering and texture generation \cite{elad2017style}. 
        Image analogy-based algorithms (e.g.,  \cite{hertzmann2001image}) examine the relationship between two input images and transfer the features to create a stylized image. However, when applied to arbitrary settings, these methods face scalability challenges. 
        Gatys et al. \cite{gatys2016image} employ a Gram matrix to extract features from a pre-trained DNN with an iterative optimization network to produce stylized images using multi-level feature correlations. Since then, many works have addressed key NST issues, including speed, control, quality, photorealism and temporal style transfer.

        As a workaround for slow iterative optimization strategies in NST, feed-forward networks are trained to minimize the same losses based on Gram matrices \cite{gatys2016image}. These feed-forward frameworks are faster than iterative optimization and appropriate for real-time deployment \cite{johnson2016perceptual}. Ulyanov et al. \cite{ulyanov2016texture} propose feed-forward network enhancements to improve example quality and variety. However, these techniques can transfer a limited number of styles because of their training process. To overcome this constraint, Dumoulin et al. \cite{dumoulin2016learned} suggest a framework based on a conditional instance normalization layer that can transfer 32 styles. Li et al. \cite{li2017diversified} propose a framework that can transfer 300 styles. However, these methods cannot transfer arbitrary styles.

        AdaIN \cite{huang2017arbitrary} 
        is a pioneering method in addressing the problem of arbitrary style transfer (AST). 
        It uses a feed-forward network to match style and content feature statistics in an intermediate layer. 
        However, AdaIN does not generalize well and faces mode collapse.  
        The WCT technique \cite{li2017universal} matches content and style covariance utilizing whitening and color transform. AdaAttN \cite{liu2021adaattn} considers both high-level and low-level features through adaptive attention normalization. The AST methods involving encoder-decoder architectures are prone to information loss due to pooling layers of the encoder network. This causes deformation of the output content.
        
    \textbf{Generative models:}
        The Variational Auto-encoder (VAE) \cite{kingma2013vae} proposes maximizing a lower bound on data probability to learn latent space representation. It learns the manifold representation of the input data distribution and generates new samples from the learned continuous latent space. 
        Despite the impressive results of GANs in image-to-image translation and style transfer, they still suffer from mode collapse, complicated training, and instability. To overcome these problems, VQ-VAE \cite{van2017neural} learns a discrete latent space instead of a continuous one. This inspired the development of VQ-GAN \cite{esser2021taming}, which uses transformers and discrete latents. Other works, such as DALL.E \cite{ramesh2021dalle} and Cogview \cite{ding2021cogview}, have also leveraged discrete latent and auto-regressive methods to achieve remarkable results in image generation. However, models based on discrete representations face challenges, such as increased accumulated errors, decreased speed for high-resolution images, and directional bias. 
        
    \textbf{Diffusion models:}
        The denoising diffusion probabilistic model is a generative model inspired by thermodynamics's ``diffusion'' process. 
        Discrete diffusion has been applied to text-generation \cite{argmaxflows} and image-generation \cite{D3PMs}, however, these models were restricted to generating 32x32 images. 
        The recently proposed VQ-diffusion \cite{gu2021vqdiff} technique enables efficient text-to-image generation and provides results comparable with the continuous paradigm. 

        \revise{Kwon et al. \cite{kwon2022diffusion} propose utilizing style and structure losses to direct the sampling process for text-guided image translation. Wang et al. \cite{wang2023stylediffusion} fine-tune diffusion models by integrating CLIP, enabling them to learn style references through text prompts. Everaert et al. \cite{everaert2023diffusion} recommend fine-tuning Stable Diffusion with a new noise distribution that mimics the style images' distribution. Diffusion-Enhanced PatchMatch \cite{hamazaspyan2023diffusion} incorporates patch-based techniques with whitening and coloring transformations in the latent space. StyleDrop \cite{sohn2024styledrop} model, which is based on the generative vision transformer Muse \cite{chang2023muse} rather than text-to-image diffusion models, produces content in diverse visual styles. Models like DreamStyler \cite{ahn2024dreamstyler} exhibit advanced textual inversion, utilizing techniques such as BLIP-2 \cite{li2023blip} and an image encoder to generate content by inverting text and content images while associating style with text. Kim et al. \cite{kim2022diffusionclip} fine-tune a pretrained DDIM to generate images based on text descriptions, introducing a local directional CLIP loss that ensures the direction between the generated image and the original image closely matches the direction between the reference (original domain) and target text (target domain). Chandramouli et al. \cite{chandramouli2022ldedit} employ a deterministic forward diffusion approach, achieving the desired manipulation by using the target text to condition the reverse diffusion process. Prompt-to-Prompt \cite{hertz2022prompt} aims to preserve some original image content by modifying the cross-attention maps. Lastly, Plug-and-Play \cite{tumanyan2023plug} investigates injecting spatial features and self-attention maps to uphold the overall structural integrity of the image. }

%% file: 20methodology.tex
\section{$D^2$Styler: A novel AST framework}
\begin{figure*}[ht]
      \centering  \includegraphics[width=1.0\textwidth]{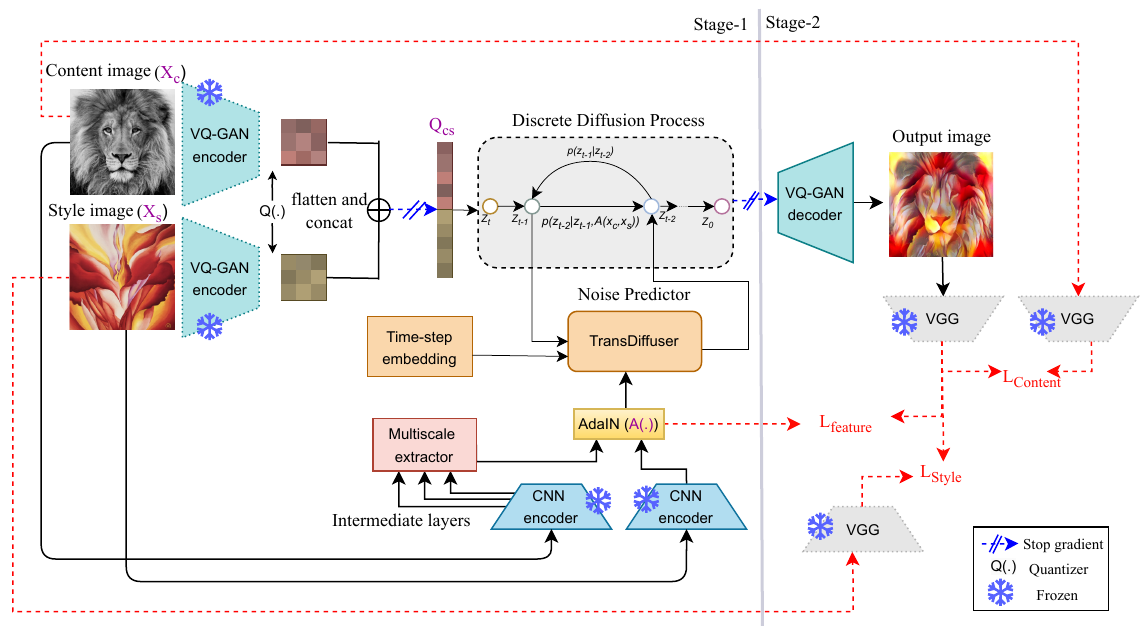}
      \caption{The architecture of the proposed method. The content and style images are encoded using a pretrained VQ-GAN encoder. The encoded input is passed through the diffusion prior conditioned on the AdaIN \cite{huang2017arbitrary} features. VQ-GAN decoder is then used to obtain the resultant image. The dotted line indicates that the diffusion prior is trained separately from the decoder.   }
      \label{fig:high_level_arch}
    \end{figure*}

$D^{2}Styler$  harnesses the power of discrete diffusion and AdaIN to address the inherent challenges of style transfer. Building upon the discrete representational capabilities of VQ-GANs, $D^{2}Styler$  pioneers a unique approach to arbitrary style transfer that promises enhanced visual quality and reduced mode collapse by intelligently navigating the latent space of content and style images.    
    Unlike traditional feed-forward methods, $D2$Styler uses the high-fidelity image generation powers of diffusion models and the fine-tuned control provided by Vector-Quantized feature spaces introduced by VQ-GAN \cite{gu2021vqdiff}. Figure \ref{fig:high_level_arch} shows the architecture of $D^2$Styler. This architecture functions in two distinct stages: stage 1 and stage 2. During stage 1, the style and content images are taken as inputs and turned into condensed features using latent discrete diffusion. Stage 2 leverages these condensed features to generate the final stylized image.

    \textbf{Stage 1:} In stage 1, both style and content images are encoded into continuous latent vectors using a VQ-GAN encoder trained on the OpenImages dataset \cite{OpenImages}. These vectors are then projected to the closest codebook item in the discrete latent space. This mapping of continuous vectors to the adjacent discrete codebook vectors is known as quantization (illustrated as Q(.) in Figure \ref{fig:high_level_arch}). The discretized nature of these vectors facilitates the grouping of similar data points, enhancing the subsequent diffusion sampling within a confined vector space. Once quantized, vectors are then flattened, concatenated, and sent to the TransDiffuser (Section \ref{TransDiffuser}). Inside the TransDiffuser, these discrete vectors go through the diffusion process, which is influenced by AdaIN features (Section \ref{sec:adain}). The outcome of this process is refined denoised features. These features proceed to the next stage (stage 2), where they play a pivotal role in reconstructing the final stylized image.

    \textbf{Stage 2:} It uses a pre-trained VQ-GAN decoder trained on OpenImages. This decoder takes improved discrete features from the TransDiffuser as input and creates a stylized picture as output. In particular, when the Stage 2 decoder is being fine-tuned, gradients from Stage 1 are blocked. This is shown in Figure \ref{fig:high_level_arch} by the blue arrow. The Stage-2 decoder incorporates a perceptual loss mechanism.
    This involves calculating the L1 distance between the hidden representation of the ground truth image and the image generated by the decoder. To do this, a pre-trained VGG model is used to derive these hidden representations for both the ground truth and the generated image. A perceptual loss term, denoted as $L_{style}$, is computed between a style image and the generated image. 
    Similarly, another term, $L_{content}$, is computed between a content image and the generated image. Together, these losses ensure that small differences at the pixel level have less effect on the network. This strikes a balance between keeping the information and sharing the desired style. This method avoids excessive stylization or under-stylization in the final image. We also compute a $L_{feature}$ loss between AdaIN features and the generated image. This loss ensures that the stylized image retains its original content while adding flair. Formulation of these losses is given in \ref{subsec:loss}
 
    \begin{figure}[ht]
            \centering
            \includegraphics[scale=0.65]{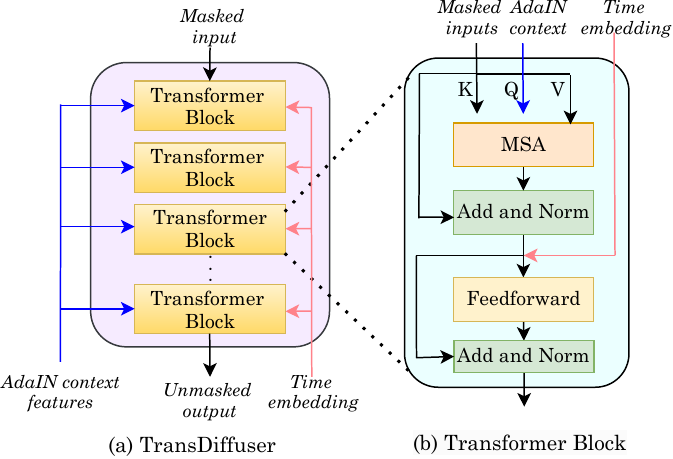}
            \caption{ (a) The proposed TransDiffuser architecture consists of transformer blocks stacked on each other. The attention query is obtained from the AdaIN block (Section  \ref{sec:adain}). 
            (b) Transformer blocks follow the traditional architecture  \cite{NIPS2017_attention_is_all_you_need} except for the querying of the AdaIN features. 
                }
            \label{fig:transdiffuser}
    \end{figure}

    \subsection{TransDiffuser}\label{TransDiffuser}
        Figure \ref{fig:transdiffuser} shows the architecture of TransDiffuser. It is designed to model discrete diffusion processes on quantized vectors. Gu et al. \cite{gu2021vqdiff} propose use of diffusion to model the discrete vector-quantized latent space of VQ-GAN. We extend this to style transfer. In TransDiffuser, we take a quantized vector ($Q_{cs}$) as input.

        During forward diffusion, this input vector ($Q_{cs}$) undergoes a gradual corruption process orchestrated by Markov chain $p(z_{t-1}|z_{t-2})$. To achieve this, tokens in $z_{t-2}$ are randomly masked. This iterative process unfolds across a fixed number of time steps ($t$), generating a sequence of latents ($z_1$, ..., $z_t$) that progressively accumulate noise. After the forward process, as visualized in Figure \ref{fig:high_level_arch}, the reverse process comes into play, starting with the noisy latent variable $z_t$. 
        This reverse process sequentially removes noise from the latent variables, eventually reconstructing the original data ($z_0$). Throughout this reverse process, AdaIN features ($A(X_c,X_s)$) are injected into each network block ($p_{\theta}(z_{t-2}|z_{t-1},A(X_c,X_s))$). These AdaIN features act as conditional cues for the diffusion process, effectively managing the equilibrium between content and style.

        To train TransDiffuser, we utilize the MLM and ELBO loss functions \cite{Devlin2019BERTPO,gu2021vqdiff}. The MLM loss facilitates the reconstruction of masked tokens, while the ELBO loss captures the probabilistic nature of content and style representations. This dual-loss approach ensures that the generated images exhibit a diverse and meaningful range of possible outcomes, striking a balance between style and content features. The resulting images maintain their original content while adopting the desired style.

        \subsection{AdaIN feature extraction}\label{sec:adain}
            Within the context of our approach to reverse diffusion, we establish the necessary conditions by harnessing both the input style image ($x_s$) and content image ($x_c$). A pivotal step involves crafting a feature map that amalgamates insights from both the style and content images. This is achieved through the utilization of pretrained CNN encoders, VGG-16, in our case. 
            To extract relevant content information, we subject the content image to the VGG encoder, extracting feature maps from specific layers such as `conv1\_2', `conv2\_2', `conv3\_2', and `conv4\_2'. We employ a multiscale extractor to adapt these extracted feature maps for compatibility with the subsequent AdaIN block. This selection of layers is particularly significant due to their ability to encapsulate high-level content details within the image \cite{gatys2016image}. 
            Likewise, the style image is also processed through the VGG encoder to extract feature maps from its final layer. These extracted feature maps from both the content and style images are then input into the AdaIN block. AdaIN aligns the statistical characteristics of the style and content features. These processed features are subsequently utilized during the reverse process in the TransDiffuser decoder, as detailed in Section \ref{TransDiffuser}. 
            Notably, the introduction of a parameter $\alpha$, in combination with the AdaIN features, enables precise control over the infusion of style into the resulting stylized image. 
            The influence of using alternative style-content CNN encoders, besides VGG-16, is explored in Section \ref{sec:ablation}.

\revised{
\subsection{Loss functions:}\label{subsec:loss} We optimized our network using a combination of four loss functions as follows.

\textit{1. Diffusion Loss ($L_{diff}$)}: 
The diffusion loss measures the ability of the model to reverse the diffusion process and generate a realistic quantized vector from noise. This loss is often implemented as a denoising score matching or noise prediction loss.
During the forward diffusion process, we diffuse the quantized vector from VQ-GAN to get $p(x_t|x_{0})$. This vector is passed through the TransDiffuser module ($P_\theta$), which is conditioned on the AdaIN features obtained from $A(X_{se}, X_{ce})$. Here, $X_{se}$ and $X_{ce}$ are the representations from the pre-trained CNN model when we pass the style $I_{s}$ and content $I_{c}$ images, respectively. At stage 1, we compute the diffusion loss as follows:

$L_{diff} = -\log(P_\theta(p(x_t | X_0), t, A(X_{se}, X_{ce})))$

In stage 2, the predicted quantized vector from $P_{\theta}$ is passed to the VQ-GAN decoder  $D$ to get the final stylized image ($\hat{x}$).  During this training stage, we block the gradient from Stage 2 to Stage 1 and only train the decoder ($D(.)$) using the following losses:

\textit{2. Style Loss ($L_{style}$)}:
Measures the difference between the style representations of the style image and the representations from the generated image through pretrained vgg-network. This ensures the output image adopts the stylistic elements (e.g., textures, colors, patterns) of the style image.

$L_{style} = \| vgg(I_s) - vgg(\hat{x}) \|$

\textit{3. Content Loss ($L_{cont}$)}: Measures the difference between the feature representations of the content image and the generated image.

$L_{cont} = \| vgg(I_c) - vgg(\hat{x}) \|$

\textit{4. Feature Loss ($L_{feat}$)}: Computes the L1 loss between the AdaIN features and the VGG representations of the generated image.  AdaIN dynamically adjusts the normalization parameters (mean and variance) of the feature maps based on the statistics of the input image. This ensures that the normalization process is tailored to the specific content and style images used. This dynamic adjustment helps better align the feature statistics of the generated image with those of the style image, leading to a more faithful style transfer.

$L_{feat} = \| A(X_{se}, X_{ce}) - vgg(\hat{x})\|$

During the training stage 2, we freeze the VGG encoder and  AdaIN A(.) encoder is unfrozen.

}

%% file: 30experimental.tex
\section{Experiments and results}\label{sec:experiments}
 \textbf{Dataset:} We select 100,000 images from the COCO dataset as content images and 78,669 images from the WikiArt dataset \cite{saleh2015large} as style images. We employ a many-to-many strategy to create 1 million content and style image pairs.      
        From these pairs, we randomly select 900,000 pairs for training and use 20,000 and 80,000 pairs for validation and testing, respectively. We employed following metrics: 
        (1) GM for faithful style adoption (2) SSIM for image similarity (3) LPIPS for perceptual resemblance and (4) PD for measuring perceptual dissimilarity.

\textbf{Experimental settings:} We trained stage-1 and stage-2 models separately on two NVIDIA A100 GPUs with 40GB VRAM each. The stage-1 is trained using the AdamW optimizer with a learning rate of 1e-5. Stage-2 is trained using the Adam optimizer with a learning rate of 1e-4. The stage-1 TransDiffuser utilizes 25 diffusion steps. The network uses six TransDiffuser blocks. 
        Training is conducted for 140,000 steps with a batch size of 32 for both the stage-1 and stage-2. Inference latency is reported on RTX 3080 Ti GPU.

%% file: 40quantResults.tex
    \subsection{Quantitative Results}\label{sec:quant}

As shown in Table \ref{table:quantativetable}, D$^2$Styler method outperforms all previous techniques on nearly all the metrics, demonstrating its robust capability in style transfer tasks.  D$^2$Styler also shows relatively small inference time.  D$^2$Styler successfully combines effectiveness and efficiency, making it a leading solution in style transfer.

        The DiT method, while slightly lagging behind D$^2$Styler in most metrics, achieves a slightly better performance in the Gram Matrix metric. This might be due to its training on a larger and more diverse dataset and its greater complexity in terms of model parameters. We have taken a pretrained DiT and finetuned it on our dataset. D$^2$Styler uses a pretrained
        VQ-GAN, however, the dataset on which DiT has been pretrained is larger than
        the one on which VQ-GAN has been pretrained.

        \begin{table}[htbp]\scriptsize
          \centering
          \caption{Quantitative results}
            \begin{tabular}{lccccc}
            \toprule
            Methods & SSIM $\uparrow$ & PD $\downarrow$ & GM $\uparrow$ & LPIPS $\downarrow$ & \revise{Inference Time (Sec)} $\downarrow$ \\
            \toprule
            CVPR'16 \cite{gatys2016image} & 0.5412 & 21.22 & 13.21 & 0.34 & \revise{13.56} \\
            \hline
            CVPR'21 \cite{kotovenko2021rethinking} & 0.6702 & 19.97 & 14.09 & 0.29 & \revise{1.72} \\
            \hline
            StyleFlow \cite{fan2022styleflow} & 0.7211 & 18.76 & 15.66 & 0.27 & \revise{1.12} \\
            \hline
            ACM MM'21 \cite{zhang2021image} & 0.7376 & 19.09 & 15.98 & 0.21 & \revise{1.42}\\
            \hline
            Cartoon-Flow (MM'22  \cite{lee2022cartoon}) & 0.7422 & 17.48 & 16.07 & 0.18 & \revise{1.37} \\
            \hline
            StyTr2 (CVPR'22 \cite{deng2022stytr2}) & 0.7778 & 16.32 & 17.46 & 0.19 & \revise{0.93} \\
            \hline
            SIGGRAPH'22 \cite{zhang2022domain} & 0.7501 & 17.37 & 16.01 & 0.17 & \revise{1.19} \\
            \hline
            InST (CVPR'23) \cite{zhang2023inversion} & 0.7347 & 18.09 & 15.21 & 0.10 & \revise{1.72} \\
            \hline
            AdaAttN (ICCV'21) \cite{liu2021adaattn} & 0.7656 & 16.77 & 17.21 & 0.09 & \textbf{\revise{0.88}} \\
            \hline
            \revise{ArtFlow (CVPR'21)} \cite{an2021artflow} & \revise{0.7512} & \revise{16.91} & \revise{17.29} & \revise{0.13} & \revise{1.18} \\
            \hline
            Stable-Diffusion (CVPR'22) \cite{rombach2022high} & 0.7789 & 17.77 & 18.21 & 0.07 & \revise{4.21} \\
            \hline
            DiT (ICCV'23) \cite{peebles2023scalable} & 0.7801 & 17.87 & \textbf{18.98} & 0.06  & \revise{5.78} \\
            \hline
            Ours w/o Encoder (Fig \ref{fig:encoder_use}(d)) & 0.7757 & 17.82 & 16.98 & 0.14 & \revise{2.85} \\
            \hline
            Ours w/o AdaIN features (Fig \ref{fig:encoder_use}(e)) &      0.7592 & 19.92 & 14.21 & 0.21 & \revise{2.54}\\
            \hline
            Ours (Fig \ref{fig:encoder_use}(c)) &       \textbf{0.7886} & \textbf{15.66} & 18.89 & \textbf{0.04} & \revise{2.92} \\
            \bottomrule
            \end{tabular}%
          \label{table:quantativetable}%
        \end{table}%

%% file: 50qualResults.tex
\subsection{Qualitative results}\label{sec:qual}

\begin{figure*}[ht]
              \includegraphics[width=0.9\textwidth]{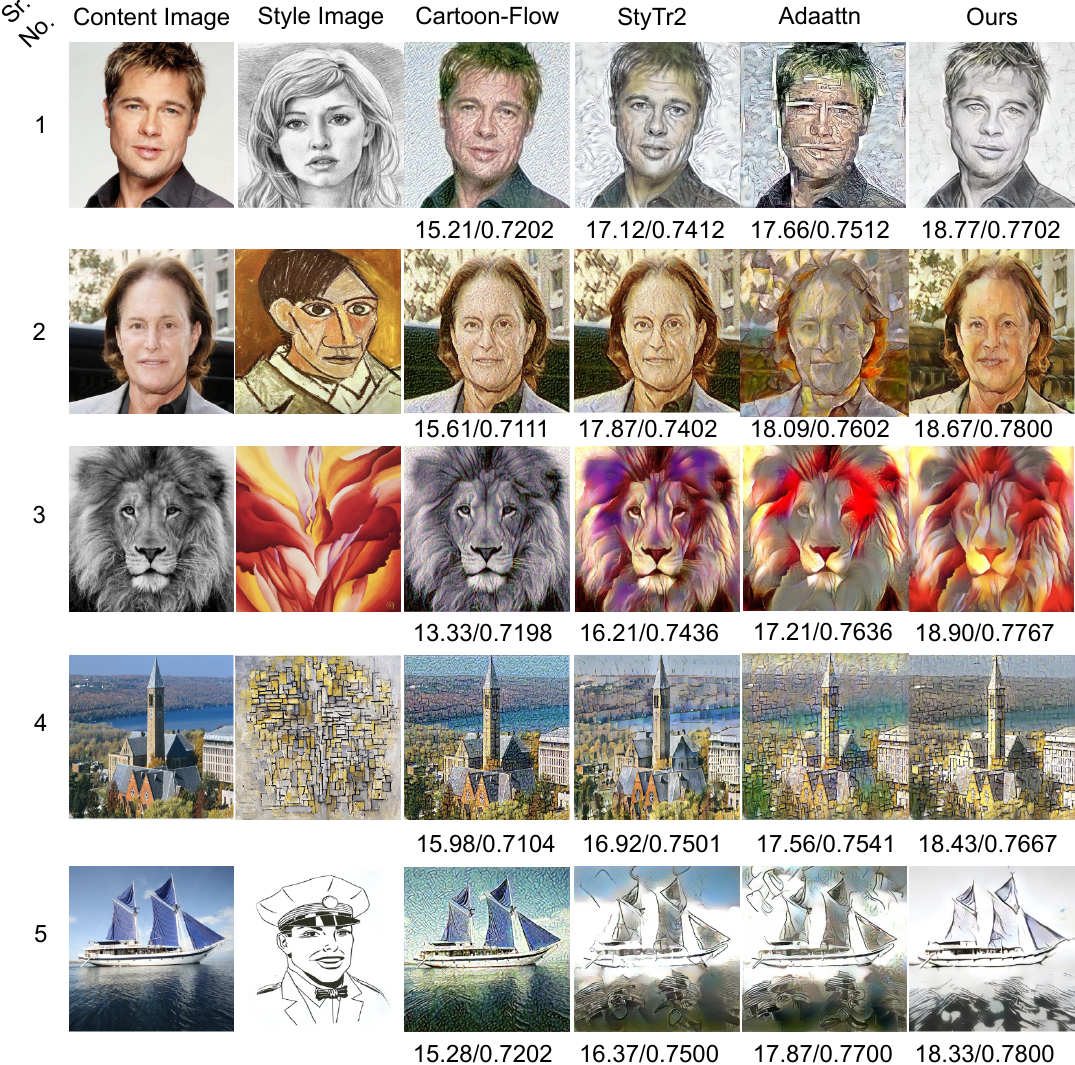}
              \caption{Qualitative results. The numbers below images show GM ($\uparrow$) and SSIM ($\uparrow$).}
              \label{fig:qualitative}
        \end{figure*}

        As depicted in Figure \ref{fig:qualitative}, D$^2$Styler effectively preserves the original content of the images while imbuing them with the desired artistic styles from the style images. The method maintains the original image’s underlying structure and details while adopting the reference image’s style, producing high-quality, visually appealing images. This capability is evident across various examples shown, where the essence and details of the original images are preserved, yet the style is convincingly and beautifully applied. This includes maintaining structural information without sacrificing the tiny details found in the style images.
     
        StyTr2 is limited in accurately mapping the reference image’s style onto the content image, producing unsatisfactory results. Cartoon-Flow, while excelling at maintaining the visual integrity of the content image, occasionally fails to adapt the reference image’s aesthetic to the output image. For instance, Cartoon-Flow can preserve the facial structure of Brad Pitt (content image) in the output image, but it cannot account for the sketch style present in the style image, hence its output image is devoid of the desired artistic effect.
        
        Both Cartoon-Flow and StyTr2 struggle to incorporate the color information present in the style image, as demonstrated in the third and fourth rows of Figure \ref{fig:qualitative}. They are unable to generate the color information of the lion images in the style image. Similarly, for the images in the fifth row, both models produce an image with the same structural quality as the content image but fail to produce the yellow box texture present in the style image on the output image.
        
        AdaAttN provides the cross attention between the style and the content features, however, it cannot properly retain similar content due to feature collapse. \revise{The use of AdaIN features as a condition to the Transdiffuser block helps in a significant boost in performance. AdaIN features are crucial for the effectiveness of style transfer methods because they separate content and style by adjusting the mean and variance of content features to match those of style features. This dynamic adjustment allows the network to adapt to a wide range of styles efficiently, enhancing the quality and visual appeal of the transfer. AdaIN simplifies the training process by focusing on normalization parameters rather than complex style representations, providing better control over the degree of stylization. Without AdaIN, style transfer methods would struggle to achieve the same level of performance, flexibility, and scalability.} From Figure \ref{fig:qualitative}, we note that for most images, AdaAttN cannot preserve the content features. In contrast, D$^2$Styler proves to be superior at capturing tiny details in the style image and accurately transferring them to the content image without sacrificing the structural information of the original image. Additionally, D$^2$Styler can robustly handle a wide range of styles and produce high-quality outputs that accurately represent the style of the reference image. Our method achieves comparable text alignment to the Diffusion-based methods for generating content images, i.e., StyTr2. This indicates that our method does not compromise the original style control capabilities of SD while learning the style of the reference images. The substantial advantage reflected in the image quality metric compared to all other methods corroborates the practicality of our approach. In summary, D$^2$Styler achieves an optimal balance between style image fidelity and content image similarity with the most pleasing image quality.

%% file: 55qualResultsSupp.tex
\begin{figure*}[ht]\centering
      \includegraphics[width=0.89\textwidth]{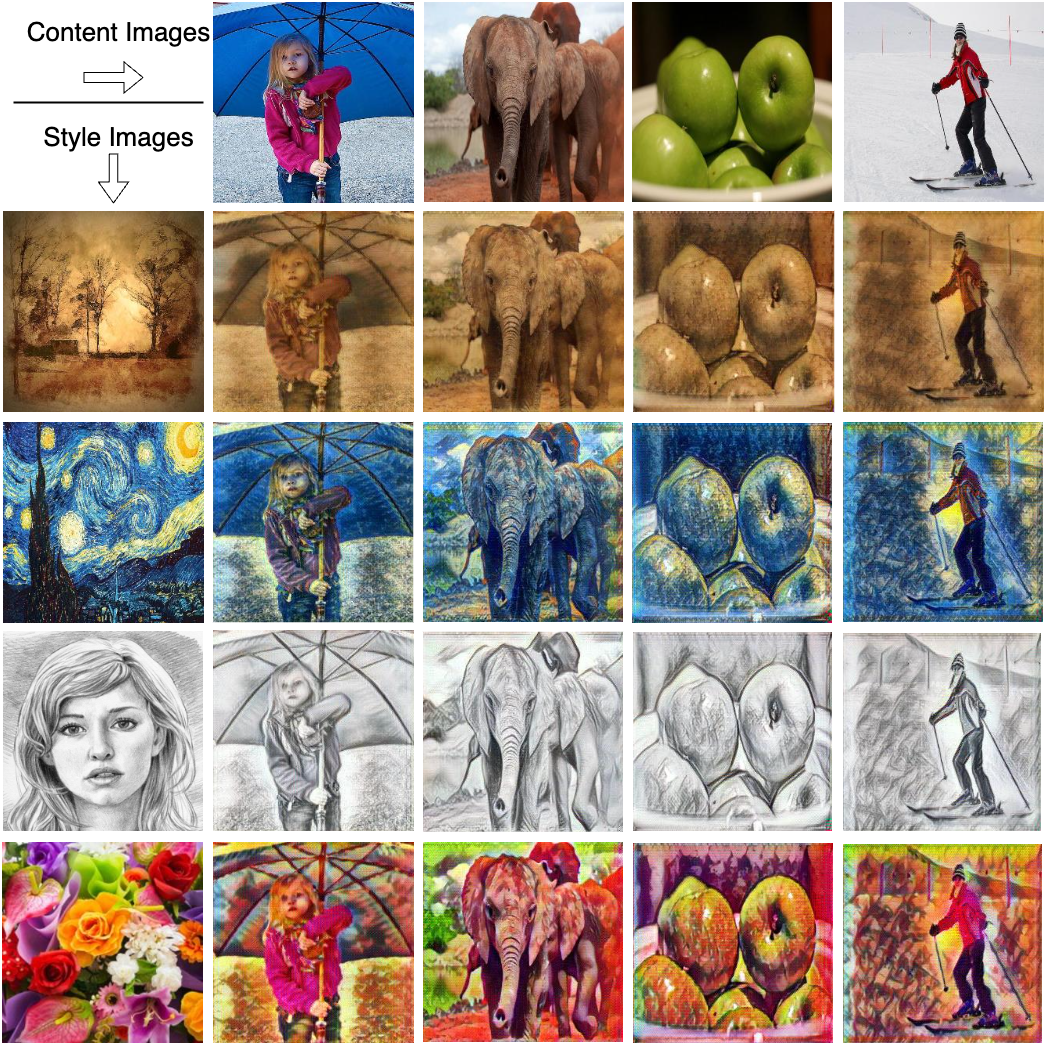}
      \caption{\revised{Qualitative results of D$^2$Styler on COCO dataset}}
      \label{fig:coc_qualitative}
\end{figure*} 
Figure \ref{fig:coc_qualitative} presents the qualitative results of D$^2$Styler on the COCO dataset, effectively demonstrating the capabilities of D$^2$Styler. 
For instance, the second row illustrates the application of a classical painting style, which imparts a rustic, textured effect reminiscent of brushstrokes and the color palette of the original artwork. Similarly, the pencil sketch style, shown in another row, transforms content images into monochromatic drawings, emphasizing lines and shading to create a hand-drawn appearance.
These results underscore the versatility and robustness of D$^2$Styler in blending various artistic styles with diverse content images, while preserving the essential characteristics of both. The model’s ability to maintain the structural integrity of the content while accurately reflecting the stylistic nuances of the artistic images demonstrates its potential applications in artistic creation and advanced image processing.

\begin{figure*}[thbp]
      \includegraphics[width=0.95\textwidth]{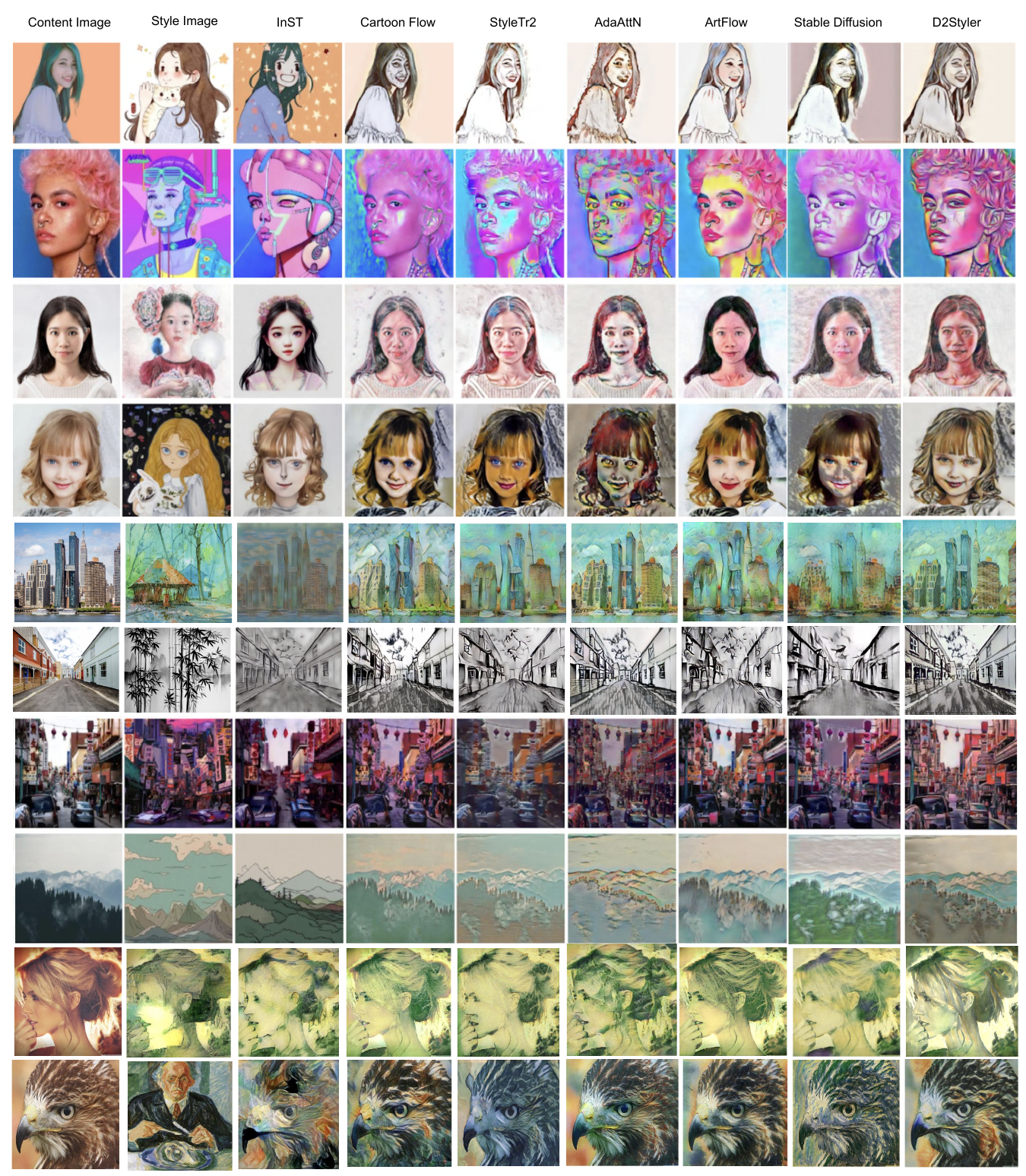}
      \caption{\revised{Qualitative comparison}} 
      \label{fig:Additionalqualitativeresults}
\end{figure*} 
\revised{Figure \ref{fig:Additionalqualitativeresults} further compares qualitative results of various techniques. InST (column \#3) \cite{zhang2023inversion} often lacks coherence in preserving the content structure. It fails to semantically transfer the color in a one-to-one correspondence. They used an additional tone transfer module \cite{huang2017arbitrary} to align the color of the content and reference images. Different methods have different preferences for retaining the colors of the content image.}

\revised{StyleTr2 (column \#5) provides reasonable style transfer results but sometimes over-emphasizes the style, leading to over-stylization. This happens because of an imbalance of content loss and style loss. If the weighting for the style loss is too high relative to the content loss, the model might prioritize transferring stylistic elements over preserving the original content structure, resulting in over-stylization. 
On the other hand, AdaAttN utilizes cross-attention mechanisms for style transfer but occasionally suffers from feature collapse, affecting content preservation (e.g., column \#6 row \#8). The per-point basis of AdaAttN leads to style degeneration, thus the stylized output is not consistent with the input reference. The content leak issue usually occurs in the stylization process because CNN-based feature representation may not sufficiently capture details in the image content. 

Furthermore, the robustness and generated visual effects of CartoonFlow (column \#4) \cite{lee2022cartoon} may degrade due to the limited capability of the feature representation. By contrast, D$^2$Styler leverages the capability of transformer-based architecture to capture long-range dependencies, hence, it significantly alleviates the content leak issue. The flow-based model has limited capability of feature representation, hence, the ArtFlow (column \#7) \cite{an2021artflow} results generally suffer from insufficient or inaccurate style. The border of stylized images may present undesirable patterns due to numerical overflow.

Stable Diffusion (column \#8) generates high-quality images, albeit with increased computational time (Table \ref{table:quantativetable}), due to the significant effort required to interpolate style and content features in its latent space to achieve perfect stylization. In contrast, D$^2$Styler demonstrates superior performance in both style fidelity and content preservation. This superiority stems from our use of diffusion in the Vector Quantized (VQ) space, where content and style features are more easily searchable and can be interpolated more effectively. By leveraging the VQ space, D$^2$Styler efficiently balances style and content, resulting in high-quality images with reduced computational complexity.}

%% file: 60versatilityResults.tex
\subsection{Versatility of D$^2$Styler}\label{sec:versatility}
\begin{figure*}[ht]
              \includegraphics[width=\textwidth]{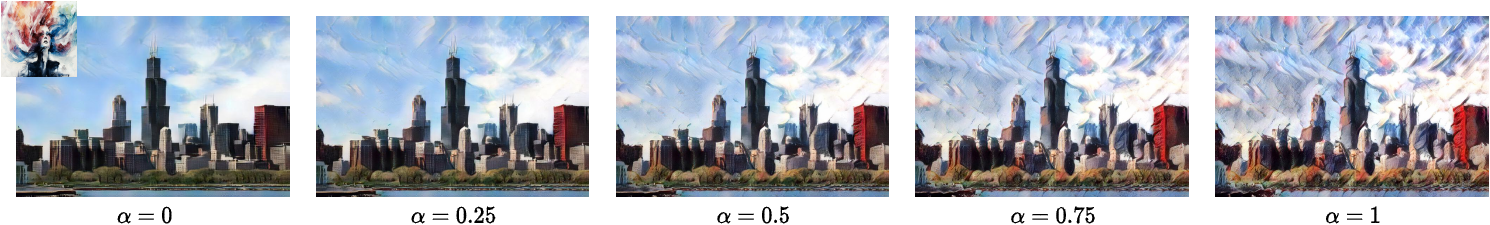}          \caption{$D^2$Styler has a parameter $\alpha$. By changing $\alpha$, the AdaIN context features can be varied to control the output image style, as shown in this image. The style image is in the inset. }
              \label{fig:control}
        \end{figure*}

\textbf{Controlling the style.} To control the amount of style output, ${D^2}$Styler introduces a weight parameter named $\alpha$ ($\alpha \in [0, 1]$) for the AdaIN context features. As shown in Figure \ref{fig:control}, the amount of style in the image is proportional to the value of the $\alpha$ parameter.
          
        \begin{figure*}
            \centering\includegraphics[width=0.9\textwidth]{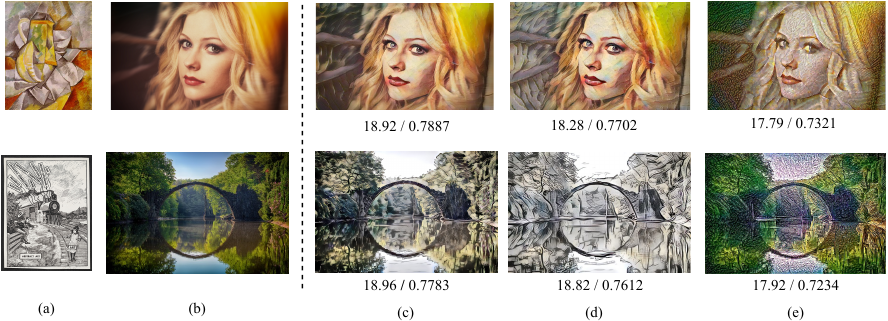}
              \caption{Input (a)  style and (b)  content images. $D^{2}$Styler results (c)  on using the encoder during inference. (d) on starting from the noise distribution during  inference. (e) without using the AdaIN features as a context to the TransDiffuser. }\label{fig:encoder_use}
        \end{figure*}

        \textbf{Use of the VQ-GAN encoder at inference.} At inference, we can use either (1) the samples from the diffusion mask prior  or (2) the VQ-GAN encoded version of the input content and style image. 
        The former gives more priority to the style texture, whereas the latter retains more content, leading to more visually pleasing images, as shown in Figure \ref{fig:encoder_use}.          
        Using the encoder also results in a slightly higher (i.e., better) SSIM metric as opposed to starting from the diffusion mask prior. Using the encoder gives a head start to the denoising process, as it contains some prior information about the content. By contrast, this prior information is absent when we start from the mask tokens themselves. 
        Table \ref{table:quantativetable} reports GM and SSIM scores for both types of inference strategies.

\begin{figure}[htbp]
                    \centering                    
                    \includegraphics[scale=0.6]{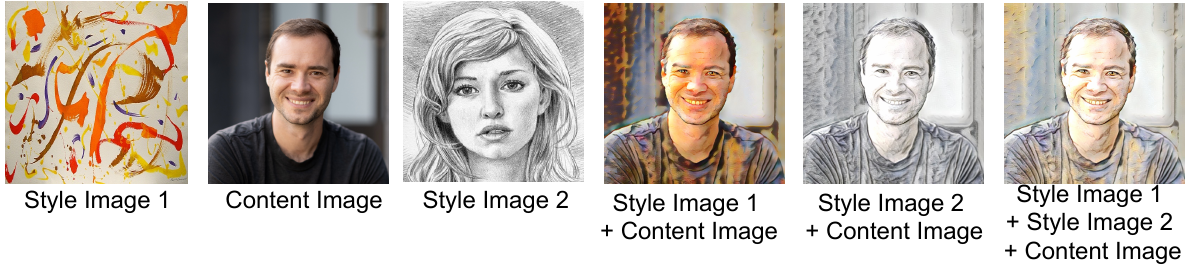}
                    \caption{Transferring the style of a single image and multiple images while preserving the content of a single content image. }            \label{fig:multi_style}
        \end{figure}

        \textbf{Multi-style transfer:} We realize multi-style transfer by passing a linear combination of  AdaIN features for each style-content combination. Each image's contribution can be controlled by using a weight for each style-content feature being combined. An example generation is shown in Figure \ref{fig:multi_style}. Evidently, the AdaIN layer serves as a context for the diffusion model, and its features can be used to control the generation of a stylized output image.
        
%% file: 80ablation.tex
\section{Ablation Study}\label{sec:ablation}
    \textbf{1. Effect of CNN encoders:} Table \ref{tab:backbone} shows the results with different CNN encoders in D$^2$Styler. Notice that using CNN encoders such as VGG, ResNet and EfficientNet leads to better results than use of transformers. NST transfers the style of a style image to a content image while keeping its spatial information. Since transformers do not incorporate locality bias inherently, they often fail to learn hierarchical representations from an input image effectively.  
    Due to this, the model faces challenges in decoupling style and content information from an image, leading to poor performance in NST.

    \begin{table}[htbp]\scriptsize
      \centering
      \caption{Ablation based on CNN Encoder}
        \begin{tabular}{p{11em}ccc}
        \toprule
         & SSIM $\uparrow$ & GM $\uparrow$ & LPIPS $\downarrow$ \\
        \toprule
        VGG-16 & \textbf{0.7886} & \textbf{18.89}  & \textbf{0.04} \\    
        Resnet-32 & 0.7721 & 16.32  & 0.08 \\    
        EfficientNet-B0 & 0.751  & 17.01  & 0.08 \\     
        ViT    & 0.7631 & 15.21  & 0.09 \\     
        PVT    & 0.7419 & 16.98  & 0.12 \\\bottomrule     
        \end{tabular}%
      \label{tab:backbone}%
    \end{table}%

\begin{figure*}[ht]
        \centering
        \includegraphics[scale=0.6]{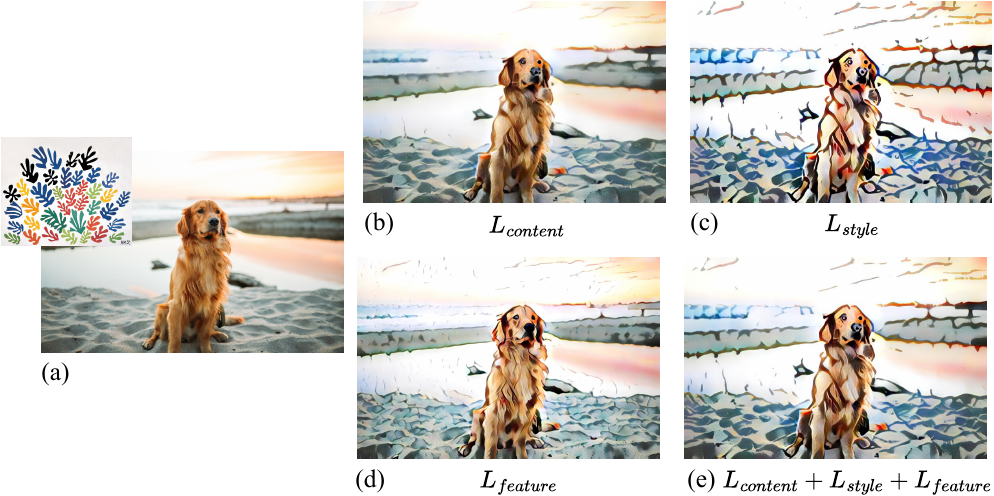}
        \caption{(a) Input style and content images. 
        $D^2$Styler results on using (b) only content loss (c)  only  style loss (d) only the proposed feature loss. (e) all three losses. }
        \label{fig:loss-ablations}
    \end{figure*}

    \textbf{2. Effect of loss functions:}
    Table \ref{tab:lossF} shows the impact of various loss functions. The best scores are obtained by using a combination of $L_{feature}$, $L_{style}$, and $L_{content}$ losses, because  this combined loss considers both style data from the encoder model and high-level features of the content. In contrast, using only $L_{style}$ (style loss) neglects content preservation, and using only $L_{content}$ (content loss) overlooks style consistency. Adding $L_{feature}$ (feature loss) to $L_{style}$ and $L_{content}$ makes the results even better by forcing a close match between the output features and the normalization data of the Adaptive Instance Normalisation (AdaIN) layer. Notably, using only $L_{feature}$ loss leads to better performance than most previous techniques \cite{kotovenko2021rethinking,gatys2016image,fan2022styleflow,zhang2021image,lee2022cartoon,zhang2022domain}. This validates the efficacy of our feature-matching approach. The above insights can be confirmed from Figure \ref{fig:loss-ablations}, which illustrates the output images with various loss functions. 
    
     \begin{table}[htbp]\scriptsize
      \centering
      \caption{Ablation based on loss functions}
        \begin{tabular}{p{15em}ccc}
        \toprule
        & SSIM $\uparrow$ &  GM  $\uparrow$ & LPIPS $\downarrow$ \\
        \toprule
        $L_{feature}$ + $L_{style}$ + $L_{content}$ & \textbf{0.7886} & \textbf{18.89}  & \textbf{0.04} \\     
        $L_{style}$ + $L_{content}$ & 0.7677 & 17.23  & 0.07 \\ 
        $L_{content}$ & 0.7501 & 16.09  & 0.09 \\ 
        $L_{style}$ & 0.7421 & 17.11  & 0.10 \\
        $L_{feature}$ & 0.7709 & 17.32  & 0.12 \\
        $L_{style} + L_{feature}$ & 0.7732 & 18.00     & 0.08 \\
        $L_{content} + L_{feature}$ & 0.7821 & 17.67  & 0.08 \\
        \bottomrule
        \end{tabular}%
      \label{tab:lossF}%
    \end{table}%
        
    \textbf{3. Effect of the number of TransDiffuser blocks:} We evaluate three versions of D$^2$Styler, namely small, medium, and large, which utilize two, four and six TransDiffuser blocks, respectively. 
    With the increasing number of blocks, the network can more accurately model the style transfer process, which improves the scores (Table \ref{tab:diffusionBlocks}). Interestingly, even small and medium versions of D$^2$Styler can outperform most previous techniques (refer Table \ref{table:quantativetable}). This proves the efficacy of our technique. 
    
    \begin{table}[htbp]\scriptsize
      \centering
      \caption{Effect of changing the number of TransDiffuser blocks}
        \begin{tabular}{p{11em}ccc}
        \toprule
         & SSIM $\uparrow$ & GM $\uparrow$ & LPIPS $\downarrow$ \\
        \toprule
        N = 6  & \textbf{0.7886} & \textbf{18.89}  & \textbf{0.04} \\     
        N = 4  & 0.7798 & 18.32  & 0.06 \\ 
        N = 2  & 0.7762 & 17.66  & 0.07 \\
        \bottomrule
        \end{tabular}%
      \label{tab:diffusionBlocks}%
    \end{table}%

\begin{figure*}[htbp]
        \centering
        \includegraphics[scale=0.32]{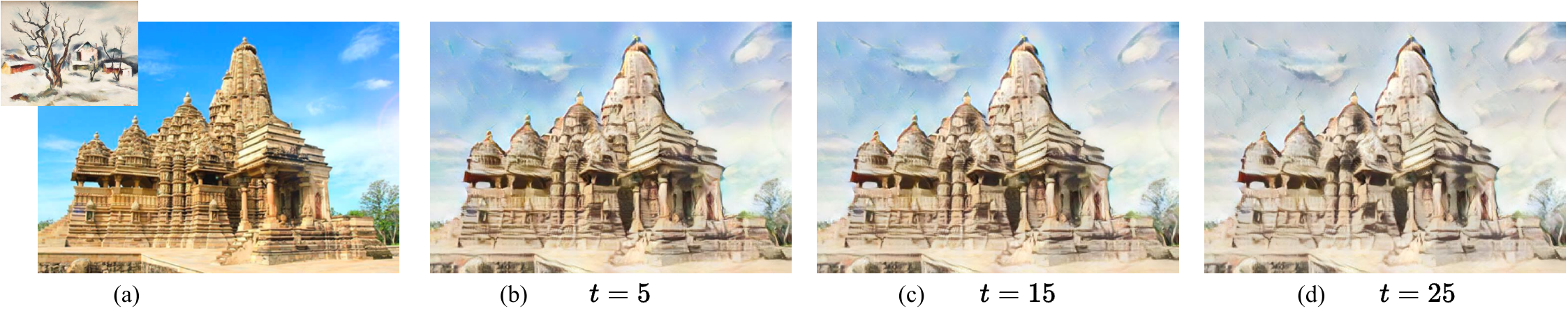}
        \caption{(a) Input style and content images. $D^{2}Styler$ results with (b) 5, (c) 15 and (d) 25 diffusion steps.           
        }
        \label{fig:t-ablation}
    \end{figure*}

    \textbf{4. Effect of changing the number of diffusion steps:}   Table  \ref{tab:diffusionSteps} shows the results with different numbers of diffusion steps.  Clearly, increasing the number of diffusion steps consistently improves image quality and structural similarity, although the returns become marginal after 25 steps. \revise{Our method demonstrates that with just 5 diffusion steps, it matches the inference times of GAN and flow-based baselines (refer to Table \ref{table:quantativetable}), and also outperforms them in all quantitative metrics. 
    This indicates our method's robustness and superior performance compared to previous baselines.} Figure \ref{fig:t-ablation} shows the output images with different numbers of diffusion steps.  
    
    \begin{table}[htbp]\scriptsize
      \centering
      \caption{Ablation Based on Diffusion Steps}      
        \begin{tabular}{ccccc}
        \toprule
         \#Steps& SSIM $\uparrow$ & GM $\uparrow$& LPIPS $\downarrow$  & \revise{Inference Time (sec)} $\downarrow$\\
        \toprule
        5      & 0.7441 & 17.02  & 0.07 & \textbf{\revise{1.93}} \\ 
        10     & 0.7512 & 17.78  & 0.06 &\revise{2.31} \\ 
        15     & 0.7663 & 18.00     & 0.06 &\revise{2.54} \\
        20     & 0.7702 & 18.88  & 0.05 &\revise{2.74} \\ 
        25     & 0.7886 & \textbf{18.89}   & \textbf{0.04} &\revise{2.92} \\ 
        50     & \textbf{0.7891} & \textbf{18.89}  & \textbf{0.04} &\revise{3.28}\\
        \bottomrule
        \end{tabular}%
      \label{tab:diffusionSteps}%
    \end{table}

%% file: 99conclusion.tex
\section{Conclusion}\label{sec:conclusion}
  We propose a novel AST technique, named D$^2$Styler, by combining the benefits of discrete diffusion with discrete representational capacity of VQ-GANs. We propose a novel way of guiding the diffusion process by incorporating  Adaptive Instance Normalisation (AdaIN) features. This allows transferring features from the style image to the content image without bias. D$^2$Styler  produces style-transferred images that are both visually appealing and accurate to the original content image in terms of their semantic significance. D$^2$Styler outperforms previous techniques and solves the problems of mode collapse, over-stylization, and under-stylization. Future work will focus on generalizing our technique to a wide range of image-processing tasks, e.g., editing and synthesis.